\documentclass[journal]{IEEEtran}
\usepackage{amsmath,amsfonts}
\usepackage{algorithmic}
\usepackage{algorithm}
\usepackage{array}
\usepackage[caption=false,font=normalsize,labelfont=sf,textfont=sf]{subfig}
\usepackage{textcomp}
\usepackage{stfloats}
\usepackage{url}
\usepackage{verbatim}
\usepackage{graphicx}
\usepackage{cite}
\usepackage{multirow}
\begin{document}

\title{Structure-Aware Graph Multi-Task Learning for Dynamic Sparse OD Demand Prediction}

\author{Ming~Xu$^{*}$ and Jiawei~Cao \thanks{$^{*}$Corresponding author: Ming Xu.} \thanks{Ming Xu and Jiawei Cao are with the Software College, Liaoning Technical University, Huludao, Liaoning 125100, China (e-mail: xum.2016@tsinghua.org.cn; 4724201684@stu.lntu.edu.cn).} \thanks{This work was supported by the Doctoral Scientific Research Foundation of Liaoning Technical University under Grant 21-1027.} \thanks{The processed datasets used in this study are available at IEEE DataPort: \protect\url{https://doi.org/10.21227/sg4n-6t97}. The source code is available at: \protect\url{https://github.com/iCityLab/SAGMTL}.} }

\maketitle
\pagestyle{plain}

\begin{abstract}
Origin–Destination (OD) demand prediction is fundamental to intelligent transportation systems, yet real-world OD flows are often dynamically sparse, long-tailed, and characterized by heterogeneous zero-flow patterns. These properties make it difficult to distinguish whether an OD connection is active from how much demand it generates once activated. Many existing methods primarily treat OD prediction as a single flow regression task, which limits their ability to model low-frequency, intermittent, and long-tailed OD interactions. To address these challenges, we propose SAGMTL, a Structure-Aware Graph Multi-Task Learning framework for dynamic sparse OD demand prediction. SAGMTL decomposes OD prediction into structural state modeling and flow intensity estimation, jointly learning regional activity states, OD connection activity, and edge-level flow intensity within a unified framework. Specifically, a node–edge collaborative representation module captures regional semantics, temporal dynamics, and spatial priors through interactive node–edge updates, producing structure-aware representations for dynamic OD interactions. Based on these representations, SAGMTL estimates OD flows by jointly modeling stable demand patterns and short-term fluctuations. A multi-constraint objective further improves sparsity awareness and structural consistency. Experiments on three real-world urban mobility datasets from Beijing, Chengdu, and Nanjing show that SAGMTL achieves superior overall performance compared with state-of-the-art baselines. Further analysis demonstrates that explicitly modeling regional activity, connection states, and flow intensity improves the robustness of dynamic sparse OD demand prediction.
\end{abstract}


\section{Introduction}
\IEEEPARstart{O}{RIGIN-DESTINATION} (OD) demand prediction is a fundamental task in intelligent transportation systems, with broad applications in traffic scheduling, capacity allocation, and urban planning \cite{ke2021od,liu2023metro,xu2019critical}. OD demand characterizes travel flows between different origin and destination regions within a specific time period \cite{xu2026rank}. Unlike node-level traffic flow prediction, OD prediction takes region pairs as the basic modeling unit and aims to capture the spatiotemporal evolution of directional travel demand \cite{zhang2021cascnn}. Therefore, OD demand prediction is inherently a structured spatiotemporal forecasting problem involving pairwise dependencies, directional interactions, and temporal dynamics. Despite its importance, real-world OD demand prediction remains challenging due to the dynamic sparsity and long-tailed distribution of OD flow data. As the number of regions increases, the number of possible OD pairs grows quadratically, i.e., $O(N^2)$, while only a small fraction of region pairs exhibit non-zero travel demand at a given temporal granularity \cite{zhuang2022uncertainty}. Moreover, travel relationships between regions are not always continuously active. Many OD pairs are intermittently activated and show non-zero flows only at a few time steps over long observation periods \cite{jiang2023tweedie}. Consequently, zero-valued observations in OD matrices do not have a unified semantic meaning. Some zero values correspond to truly nonexistent travel relationships, some indicate temporary inactivity of potentially valid OD connections, and others reflect low-frequency or weakly observed travel relationships. This heterogeneous generation mechanism of zero observations introduces substantial structural state heterogeneity into OD data, making it difficult to distinguish whether an OD pair is structurally disconnected, temporarily inactive, or latently connected.

Existing OD prediction methods usually formulate OD demand forecasting as a direct regression problem over all OD pairs. Under highly sparse observations, such a formulation tends to treat zero values as homogeneous flow targets and overlooks the latent structural states behind them. As a result, model learning can be dominated by a large number of zero entries, which weakens the ability to identify potentially active or low-frequency OD relationships. Although previous studies have attempted to alleviate OD sparsity through zero-inflated modeling, long-tailed distribution fitting, and uncertainty estimation \cite{zhuang2022uncertainty,jiang2023tweedie}, most of them mainly address sparsity at the distributional or loss-design level. The structural distinction between whether an OD connection exists and how strong the corresponding travel demand is remains insufficiently modeled. Therefore, explicitly decomposing OD prediction into connectivity state recognition and conditional flow intensity estimation is essential for improving prediction performance under dynamic sparse observations.

In addition to historical OD flows, the formation of OD demand is closely related to regional functional attributes, population distribution, land-use patterns, and urban activity structures \cite{rong2024generate,zhao2025heterogeneous}. Different regions may serve heterogeneous urban functions, such as residential areas, employment centers, commercial districts, and transportation hubs, leading to significant differences in travel generation and attraction capacities \cite{lin2019deepstn}. As a result, OD relationships are shaped not only by geographical proximity but also by functional complementarity and non-local semantic interactions \cite{yao2019stdn}. This is particularly important for low-frequency and intermittent OD pairs, whose latent connectivity is difficult to infer from historical flow observations alone. Stable regional semantic information can provide useful priors for identifying potential travel relationships and enhancing the robustness of OD prediction under sparse and long-tailed distributions.

Based on the above analysis, the key challenge in OD demand prediction lies in jointly modeling the connectivity states and flow intensities of OD relationships under dynamic sparse observations, while incorporating regional semantic information to infer low-frequency and latent OD patterns. To address this challenge, we propose SAGMTL, a structure-aware graph multi-task learning framework for dynamic sparse OD demand prediction. SAGMTL consists of three tightly coupled components: a representation learning module, a multi-task decoding module, and a constraint-driven optimization module. Specifically, the representation learning module jointly encodes static regional semantics, dynamic temporal states, OD interaction patterns, and spatial priors into unified node and edge representations. Based on these representations, the multi-task decoding module jointly predicts OD connectivity states and conditional flow intensities, enabling the model to distinguish structural existence from demand magnitude. Furthermore, the constraint-driven optimization module introduces sparsity control, temporal consistency, and marginal flow consistency constraints to improve robustness under dynamic sparse and long-tailed OD distributions.

Experimental results on three real-world taxi and ride-hailing OD datasets from Beijing, Chengdu, and Nanjing demonstrate that the proposed SAGMTL consistently outperforms a variety of classical spatiotemporal forecasting models and advanced OD prediction methods. The results show that SAGMTL achieves superior performance in both flow prediction accuracy and connectivity state identification, validating its effectiveness and robustness for dynamic sparse OD demand prediction.

In summary, the main contributions of this paper are as follows:
\begin{itemize}
    \item We propose a structure-aware multi-task learning framework for dynamic sparse OD demand prediction. The framework reformulates OD forecasting as a joint learning problem of connectivity state recognition and conditional flow intensity estimation, explicitly decomposing OD relationships into structural existence and demand magnitude.

    \item We design a node--edge collaborative representation learning mechanism that integrates regional functional semantics, temporal dynamics, OD interaction patterns, and spatial priors through interactive node--edge updates. This mechanism enhances the model's ability to infer low-frequency and intermittent OD relationships under sparse observations.

    \item We introduce a multi-constraint optimization objective tailored to dynamic sparse and long-tailed OD distributions. By jointly incorporating flow fitting, structural supervision, sparsity regularization, temporal consistency, and marginal flow consistency, the proposed objective improves prediction robustness and stability under severe sparsity and temporal fluctuations.
\end{itemize}

\section{ RELATED WORK}

\subsection{OD Demand Prediction}

Early OD demand prediction methods model travel demand between region pairs using matrix-, tensor-, or graph-based representations, where OD flows are treated as the fundamental forecasting units. Representative deep learning approaches focus on capturing spatiotemporal dependencies within OD matrices \cite{yuan2025prototype,gong2026dstcn,zhang2025gnnod}. CSTN \cite{liu2019cstn} models OD demand by integrating spatial, temporal, and global contextual information, while CAS-CNN \cite{zhang2021cascnn} alleviates sparsity and high dimensionality in metro OD prediction via channel-attention-based convolutional structures. Other studies explore partially observed OD estimation from inflow/outflow signals \cite{cheng2022metro} and incorporate external contextual factors such as land use \cite{shanthappa2024public}. With the development of graph neural networks, MT-MF-GCN \cite{feng2022mtmfgcn} jointly models region-level and OD-level tasks, while GEML \cite{wang2019geml} and MPGCN \cite{shi2020mpgcn} learn spatial dependencies from geographic, semantic, and directional graph structures. Hybrid spatiotemporal models further combine graph-based spatial learning with sequential temporal modeling for OD prediction tasks \cite{han2022continuous,zhang2022dynamic,rong2025multihop}. However, existing methods predominantly treat OD prediction as a flow regression problem, focusing on demand magnitude estimation while ignoring the structural nature of OD relationships.

\subsection{Dynamic Sparse OD and Structural Gap}

Real-world OD data exhibit severe dynamic sparsity and long-tailed distributions, where only a small fraction of region pairs are active at each time step despite a quadratic growth in potential connections. Moreover, OD interactions are highly intermittent, leading to complex zero-heavy and imbalanced observations. To address sparsity, existing studies mainly focus on probabilistic modeling and uncertainty-aware learning. For example, Zhuang et al.~\cite{zhuang2022uncertainty} estimate uncertainty in sparse travel demand using spatiotemporal graph models, while Jiang et al.~\cite{jiang2023tweedie} adopt Tweedie-based distributions to handle zero-inflated and long-tailed OD flows. Although these approaches improve robustness under sparse observations, they still model OD prediction as a pure flow estimation problem without explicitly distinguishing underlying structural states. Recent advances in dynamic graph learning suggest that edge evolution should be explicitly modeled rather than implicitly inferred from node dynamics. Temporal graph models and streaming link prediction methods \cite{xu2020tgat,sankar2020dysat} demonstrate the importance of capturing dynamic connectivity explicitly. Works such as TGAE \cite{wang2023tgae} model OD networks as evolving weighted graphs, while streaming temporal link prediction \cite{yang2026tlp} treats OD interactions as time-dependent edges. These studies indicate that OD connectivity is not merely a byproduct of flow prediction but a fundamental predictive target. Additionally, structural consistency constraints such as flow conservation have been explored to improve OD estimation quality \cite{zhang2026conservation}, further highlighting the importance of structural regularization.

Overall, existing studies have explored sparse flow modeling, probabilistic estimation, and structural constraints separately. However, they lack a unified formulation that jointly models node activity, connection existence, and flow intensity, leaving their hierarchical dependencies underexplored.

\subsection{ Spatio-Temporal Graph Modeling}

Graph-based spatiotemporal modeling has become a dominant paradigm in traffic forecasting due to its ability to capture non-Euclidean spatial dependencies \cite{shao2022stid,jiang2022survey,jin2024survey}. Early models such as DCRNN \cite{li2018dcrnn}, STGCN \cite{yu2018stgcn}, and T-GCN \cite{zhao2020tgcn} integrate graph convolution with temporal modules to model spatial diffusion and temporal evolution over predefined graphs. To overcome limitations of static graphs, adaptive and dynamic graph learning methods such as Graph WaveNet \cite{wu2019graphwavenet}, AGCRN \cite{bai2020agcrn}, and MTGNN \cite{wu2020mtgnn} learn latent spatial dependencies via adaptive adjacency matrices. Continuous and dynamic formulations such as STGODE \cite{fang2021stgode} and D2STGNN \cite{shao2022d2stgnn} further enhance long-range dependency modeling. Transformer-based approaches \cite{guo2019astgcn,zheng2020gman,jiang2023pdformer,liu2023staeformer} improve global spatiotemporal modeling, while PatchSTG \cite{fang2025patchstg} improves efficiency via spatial patching mechanisms. However, most graph-based models remain node-centric, where edges serve as information propagation channels rather than prediction targets. In contrast, OD prediction fundamentally requires modeling edge-level existence and flow intensity dynamics. Motivated by this limitation, this work proposes a node--edge collaborative learning framework that jointly models node activity, OD connection existence, and conditional flow intensity within a unified multi-task paradigm, enabling effective learning under dynamic sparse and long-tailed OD distributions.

\section{ METHODOLOGY}

\subsection{Problem Formulation}
At time step $t$, the OD system is represented as a directed weighted graph
\begin{equation}
G_t = (V, E, \mathbf{Y}_t, \mathbf{Z}_t^{e})
\label{eq}
\end{equation}

where $V=\{v_1,v_2,\ldots,v_N\}$ denotes the set of nodes, and each node represents a traffic analysis zone that can act as both an origin and a destination. $E\subseteq V\times V$ denotes the OD edge set constructed from historical observations, containing region pairs with at least one recorded interaction. In this work, both $V$ and $E$ are assumed to be fixed during inference, and we focus on modeling temporal variations over existing OD pairs without considering cold-start edges.

For each edge $(v_i,v_j)\in E$, the observed OD flow at time $t$ is denoted as $y_{ij,t}$. We define its binary activation state as $z_{ij,t}^{e}=\mathbb{I}(y_{ij,t}>0)$, where $z_{ij,t}^{e}=1$ indicates that the OD pair is active at time step $t$, and $z_{ij,t}^{e}=0$ otherwise. Accordingly, the flow matrix $\mathbf{Y}_t\in\mathbb{R}_{\geq 0}^{|E|}$ and the edge state vector $\mathbf{Z}_t^{e}\in\{0,1\}^{|E|}$ are obtained by stacking all edges in $E$. Each node $v_i$ is associated with a static feature vector $\mathbf{x}_i^s$ and a dynamic feature vector $\mathbf{x}_{i,t}^{d}$. The static features describe time-invariant regional attributes, while the dynamic features characterize the recent travel activity of each region. Specifically, the dynamic node feature is defined as:

\begin{equation}
\mathbf{x}_{i,t}^{d}
=
\left[
deg_{i,t}^{out},
deg_{i,t}^{in},
flow_{i,t}^{out},
flow_{i,t}^{in}
\right]
\label{eq:dynamic_node_feature}
\end{equation}

where
\begin{equation}
deg_{i,t}^{out}
=
\sum_{j:(v_i,v_j)\in E} z_{ij,t}^{e}
\label{eq:dynamic_out_degree}
\end{equation}
\begin{equation}
deg_{i,t}^{in}
=
\sum_{j:(v_j,v_i)\in E} z_{ji,t}^{e}
\label{eq:dynamic_in_degree}
\end{equation}
\begin{equation}
flow_{i,t}^{out}
=
\sum_{j:(v_i,v_j)\in E} y_{ij,t}
\label{eq:dynamic_out_flow}
\end{equation}
\begin{equation}
flow_{i,t}^{in}
=
\sum_{j:(v_j,v_i)\in E} y_{ji,t}
\label{eq:dynamic_in_flow}
\end{equation}

Given OD observations $\{{G}_{t-T+1},\ldots,{G}_{t}\}$, node static features ${X}^s$, node dynamic feature sequence $\{{X}_{t-T+1}^{d},\ldots,{X}_{t}^{d}\}$, and spatial distance matrix $\mathbf{D}\in\mathbb{R}^{|V|\times |V|}$, the primary objective is to predict future OD flow intensities over the next $K$ time steps:
\begin{equation}
\{\hat{\mathbf{Y}}_{t+1},\ldots,\hat{\mathbf{Y}}_{t+K}\}
\label{eq:flow_prediction_task}
\end{equation}

To better capture the dynamics of sparsity in OD interactions, we further introduce two auxiliary tasks. The first is edge activity prediction, which estimates whether each OD pair will be active:

\begin{equation}
    \{\hat{Z}_{t+1}^{e},\ldots,\hat{Z}_{t+K}^{e}\}, \quad \hat{Z}_{t+k}^{e}\in[0,1]^{|E|} \label{eq:edge_activity_task} \end{equation} 
    
The second is node activity prediction, which models whether each region will be active as an origin or as a destination:

 \begin{equation} 
 \{\hat{Z}_{t+1}^{n},\ldots,\hat{Z}_{t+K}^{n}\}, \quad \hat{Z}_{t+k}^{n}\in[0,1]^{|V|\times 2} \label{eq:node_activity_task}
\end{equation}

For node $v_i$, the two dimensions of $\hat{\mathbf{Z}}_{t+k}^{n}$ correspond to origin activity and destination activity, respectively. The ground-truth node activity labels are derived from edge activity states:
\begin{equation}
z_{i,t}^{n,out}
=
\mathbb{I}
\left(
\sum_{j:(v_i,v_j)\in E} z_{ij,t}^{e} > 0
\right)
\end{equation}
\begin{equation}
z_{i,t}^{n,in}
=
\mathbb{I}
\left(
\sum_{j:(v_j,v_i)\in E} z_{ji,t}^{e} > 0
\right)
\end{equation}

Through this formulation, sparse OD demand prediction is decomposed into three coupled subproblems: regional activity modeling, OD connection state prediction, and flow intensity estimation.

\subsection{ Method Overview}

\begin{figure*}[!t]
\centering
\includegraphics[width=0.98\textwidth]{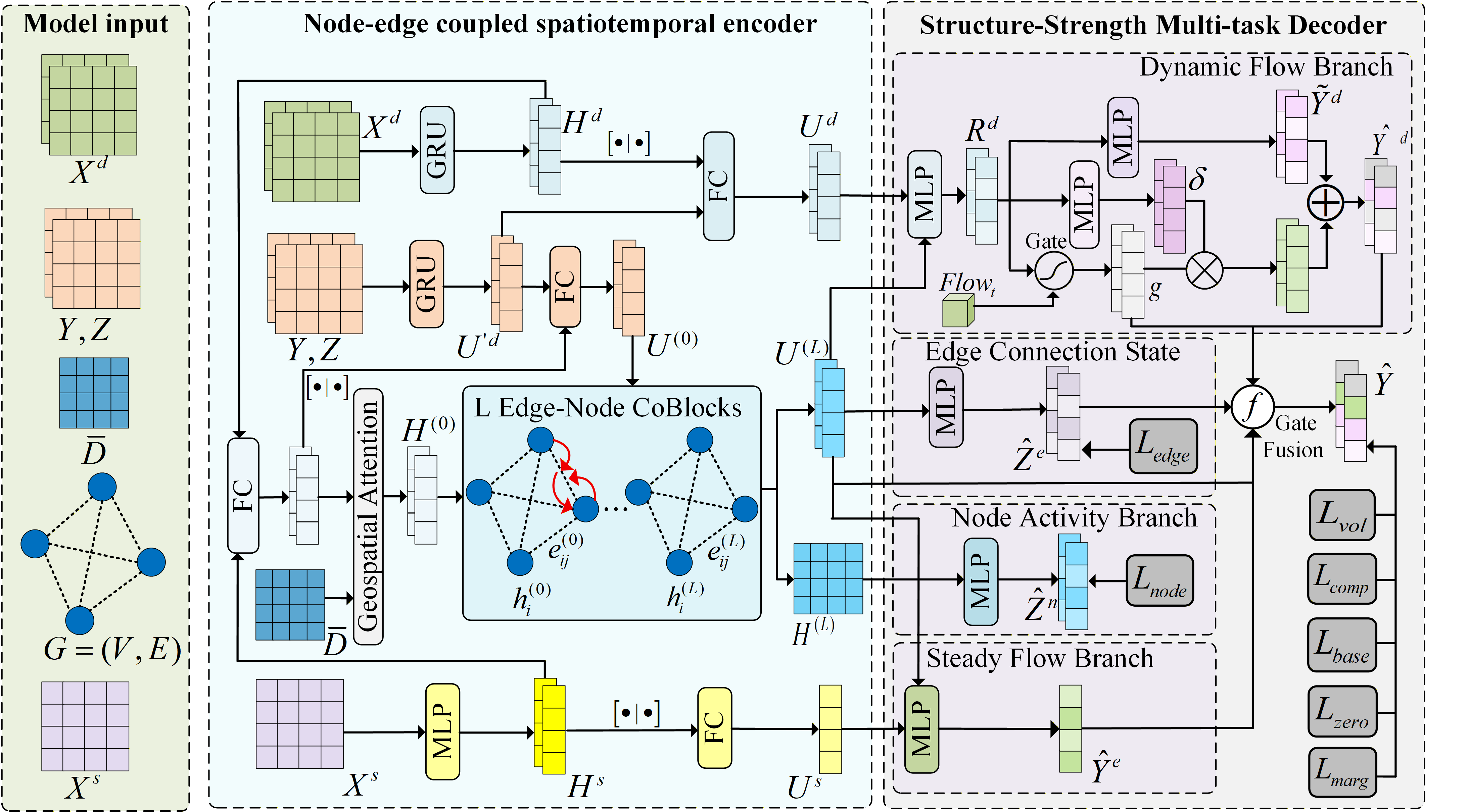}
\caption{Overall architecture of the proposed SAGMTL framework. Given historical OD flows, edge activation states, spatial relationships, and static regional attributes, SAGMTL first learns structure-aware node--edge representations through the node--edge coupled spatiotemporal encoder. The structure--strength multi-task decoder consists of four branches: the steady flow branch models stable OD demand patterns, the dynamic flow branch captures short-term residual variations, the edge connection state branch predicts OD edge activation, and the node activity branch estimates origin and destination activity states. The steady and dynamic flow branches are further fused to generate final OD flow predictions, while structural supervision and multiple regularization losses are jointly optimized to improve dynamic sparse OD demand prediction.}
\label{fig:framework}
\end{figure*}

As illustrated in Fig.~\ref{fig:framework}, SAGMTL adopts an encoder--decoder architecture consisting of a representation learning module, a multi-task decoding module, and a joint optimization module. Given historical OD observations, regional attributes, and spatial relationships, the representation learning module learns unified node--edge representations that encode regional semantics, temporal dynamics, and structural dependencies. Based on the learned representations, the multi-task decoder jointly predicts node activity states, OD connection states, and future flow intensities \cite{caruana1997multitask}. To improve robustness under dynamically sparse OD distributions, the joint optimization module incorporates flow supervision, structural supervision, and consistency regularization during training. The model ultimately outputs the predicted OD flows and corresponding connection states over the next $K$ time steps.

For notational simplicity, $\mathrm{MLP}(\cdot)$ denotes a generic multilayer perceptron. Outputs representing probabilities are mapped to $[0,1]$ via a sigmoid function, while nonnegative flow variables are generated using a softplus activation. Unless otherwise specified, hidden representations and intermediate variables are unconstrained.

\subsection{Node--Edge Coupled Spatio-temporal Encoder}

\subsubsection{Initial Node Representation}

For each node $v_i$, its static functional attribute vector $\mathbf{x}_i^s$ is mapped into a static embedding $\mathbf{h}_i^s$ through a linear transformation. To capture temporal variations in regional activity, a GRU is used to encode the dynamic node feature sequence over the past $T$ time steps:

\begin{equation}
h_{i,t}^{d} = \mathrm{GRU}_{n} \left( x_{i,t-T+1}^{d},\ldots,x_{i,t}^{d} \right) \label{eq:node_dynamic_gru} 
\end{equation}

The static semantic embedding and dynamic state representation are then fused to obtain the initial node representation:
\begin{equation}
h_{i,t}^{(0)}
=
\mathrm{MLP}_{h}
\left(
h_i^{s} \Vert h_{i,t}^{d}
\right)
\label{eq:initial_node_rep}
\end{equation}

where the operator $\Vert$ denotes vector concatenation.

To incorporate spatial priors, we further introduce a distance-aware spatial attention mechanism. The geographical distance between nodes is first transformed as:
\begin{equation} 
\bar{D}_{ij} = \log\left(1+\frac{D_{ij}}{d_0}\right) \label{eq:spatial_distance_transform} 
\end{equation}

where $d_0$ is a distance normalization factor. Let $\mathbf{H}_t^{(0)}=\left[\mathbf{h}_{1,t}^{(0)},\ldots,\mathbf{h}_{|V|,t}^{(0)}\right]^{\top}$. The query, key, and value matrices are computed as:
\begin{equation}
\mathbf{Q}_t=\mathbf{H}_t^{(0)}\mathbf{W}_Q,\quad
\mathbf{K}_t=\mathbf{H}_t^{(0)}\mathbf{W}_K,\quad
\mathbf{V}_t=\mathbf{H}_t^{(0)}\mathbf{W}_V
\label{eq:qkv_projection}
\end{equation}

The distance-aware spatial representation is obtained by: 
\begin{equation}
H_t^{p} = \mathrm{softmax} \left( \frac{Q_t K_t^{\top}}{\sqrt{d_{\alpha}}} - \gamma \bar{D} + M \right) V_t \label{eq:spatial_attention} 
\end{equation}

where $\gamma=\mathrm{softplus}(\theta_{\gamma})$ is a learnable non-negative distance decay coefficient, $M$ is the self-connection mask, where $M_{ij}=-\infty$ if $i=j$ and $M_{ij}=0$ otherwise. 

The spatially enhanced node representation is obtained through a residual connection:
\begin{equation} 
{H}_t^{(0)} = H_t^{(0)} + \lambda_p H_t^{p} \label{eq:spatial_residual} 
\end{equation} 

where $\lambda_p$ controls the contribution of the spatial attention representation. 

\subsubsection{Initial Edge Representation}

For each OD edge $(v_i,v_j)$, the static edge representation is constructed from the semantic embeddings of its origin and destination nodes : 
\begin{equation} 
u_{ij}^{s} = \mathrm{MLP}_{s}^{enc} \left( h_i^{s} \Vert h_j^{s} \right) \label{eq:static_edge_rep} 
\end{equation} 

This representation captures relatively stable OD interaction tendencies induced by regional functional semantics.

To model the short-term OD interaction dynamics, an edge-intrinsic dynamic representation is learned for each OD edge by applying a GRU to its recent flow and connection state sequence: 
\begin{equation}
u_{ij,t}^{\prime d}
=
\mathrm{GRU}_{e}
\left(
\left[\mathbf{y}_{ij}^{t-T+1}, \mathbf{z}_{ij}^{e,t-T+1}\right],
\ldots,
\left[\mathbf{y}_{ij}^{t}, \mathbf{z}_{ij}^{e,t}\right]
\right)
\label{eq:dynamic_edge_gru}
\end{equation}

This edge-intrinsic dynamic representation serves two purposes. It is fused with the dynamic representations of the origin and destination nodes to obtain an endpoint-enhanced task-oriented dynamic edge representation: 
\begin{equation} 
u_{ij,t}^{d} = \mathrm{MLP}_{u}^{enc} \left( u_{ij,t}^{\prime d} \Vert h_{i,t}^{d} \Vert h_{j,t}^{d} \right) \label{eq:edge_dynamic_fusion} 
\end{equation}

It is also combined with the initial representations of the endpoint nodes to construct the initial edge representation for subsequent node–edge collaborative updates: 
\begin{equation}
u_{ij,t}^{(0)}
=
\mathrm{MLP}_{u}^{init}
\left(
u_{ij,t}^{\prime d}
\Vert h_{i,t}^{(0)}
\Vert h_{j,t}^{(0)}
\right)
\label{eq:initial_edge_rep}
\end{equation}

\subsubsection{Node--Edge Collaborative Update} 

Based on the initial node and edge representations, SAGMTL performs node--edge collaborative updates to model the bidirectional dependencies between regional states and OD interactions \cite{gilmer2017nmp}. At the $l$-th update layer, the node and edge representations are denoted as $h_{i,t}^{(l)}$ and $u_{ij,t}^{(l)}$, respectively.

First, edge-to-node message aggregation is performed. For node $v_i$, outgoing and incoming edge messages are aggregated separately:
\begin{align} 
m_{i,t}^{\mathrm{out},(l)} &= \frac{1}{c_{i}^{\mathrm{out}}} \sum_{j:(v_i,v_j)\in E} W^{\mathrm{out}}u_{ij,t}^{(l)} \\ m_{i,t}^{\mathrm{in},(l)} &= \frac{1}{c_{i}^{\mathrm{in}}} \sum_{j:(v_j,v_i)\in E} W^{\mathrm{in}}u_{ji,t}^{(l)} \label{eq:edge_to_node_msg} 
\end{align} 

Here, $W^{\mathrm{out}}$ and $W^{\mathrm{in}}$ are learnable transformation matrices. $c_{i}^{\mathrm{out}}$ and $c_{i}^{\mathrm{in}}$ denote the static out-degree and static in-degree of node $v_i$ over the edge set $E$, respectively. The node representation is updated as: 
\begin{equation} 
h_{i,t}^{(l+1)} = h_{i,t}^{(l)} + \mathrm{MLP}_{h}^{\mathrm{upd}} \left( h_{i,t}^{(l)} \Vert m_{i,t}^{\mathrm{out},(l)} \Vert m_{i,t}^{\mathrm{in},(l)} \right) \label{eq:node_update} 
\end{equation} 

Then, node-to-edge information propagation is performed. For each OD edge  $(v_i,v_j)$, the edge representation is updated as: 
\begin{equation} 
\begin{aligned} u_{ij,t}^{(l+1)} &= u_{ij,t}^{(l)} + \mathrm{MLP}_{u}^{\mathrm{upd}} \Big( u_{ij,t}^{(l)} \Vert h_{i,t}^{(l+1)} \Vert h_{j,t}^{(l+1)} \\ &\qquad\qquad \Vert \left|h_{i,t}^{(l+1)}-h_{j,t}^{(l+1)}\right| \Big) \end{aligned} \label{eq:edge_update} 
\end{equation}

Through alternating edge-to-node and node-to-edge propagation, the encoder captures both the influence of regional activity patterns on OD connections and the feedback effect of OD interaction on regional states.

After $L$ collaborative update layers, the encoder outputs the high-order node representation set:
\begin{equation}
H_t^{(L)}
=
\left\{
h_{i,t}^{(L)}
\mid
v_i \in V
\right\}
\label{eq:high_order_node_set}
\end{equation}

and the high-order edge representation set:
\begin{equation}
U_t^{(L)}
=
\left\{
u_{ij,t}^{(L)}
\mid
(v_i,v_j)\in E
\right\}
\label{eq:high_order_edge_set}
\end{equation}

The static edge representation set
$U^{s}=\{u_{ij}^{s}\mid (v_i,v_j)\in E\}$
and the endpoint-enhanced dynamic edge representation set
$U^{d}=\{u_{ij,t}^{d}\mid (v_i,v_j)\in E\}$
are also retained for subsequent multitask decoding.

\subsection{Connection--Flow Decomposed Multitask Decoder}

Based on the learned node and edge representations, SAGMTL employs a connection--flow decomposed multitask decoder to jointly predict OD connection states and future flow intensities. The decoder consists of four branches: an edge connection state branch, a node activity branch, a steady flow branch, and a dynamic flow branch. The final OD flow is generated by a connection-guided gated fusion mechanism.

\subsubsection{Edge Connection State Branch}

For each OD edge $(v_i,v_j)$, the edge connection state branch predicts its activation probabilities over the next $K$ time steps:
\begin{equation}
\left[
\hat{z}_{ij,t+1}^{e},\ldots,\hat{z}_{ij,t+K}^{e}
\right]
=
\mathrm{MLP}_{e}
\left(
u_{ij,t}^{(L)}
\right)
\label{eq:edge_activation_prediction}
\end{equation}

Here, $\hat{z}_{ij,t+k}^{e}\in[0,1]$ denotes the probability that edge $(v_i,v_j)$ will be activated at future time step $t+k$. This branch explicitly models OD connection states and provides structural information for flow intensity estimation.

\subsubsection{Node Activity Branch}

To provide region-level structural supervision, SAGMTL further predicts whether each node will be active as an origin or as a destination. For node $v_i$, the node activity branch is defined as:
\begin{equation}
\left[
\hat{z}_{i,t+1}^{n,\mathrm{out}},
\hat{z}_{i,t+1}^{n,\mathrm{in}},
\ldots,
\hat{z}_{i,t+K}^{n,\mathrm{out}},
\hat{z}_{i,t+K}^{n,\mathrm{in}}
\right]
=
\mathrm{MLP}_{n}
\left(
h_{i,t}^{(L)}
\right)
\label{eq:node_activity_prediction}
\end{equation}

For each future time step $t+k$, the two dimensions correspond to $\hat{z}_{i,t+k}^{n,out}$ and $\hat{z}_{i,t+k}^{n,in}$, representing the probabilities that node $v_i$ is active as an origin and as a destination, respectively. This auxiliary task helps the model learn regional travel generation and attraction patterns, which are especially useful under sparse OD observations.

\subsubsection{Steady Flow Branch}

OD flow usually contains relatively stable patterns induced by regional functions, spatial relationships, and long-term travel regularities. To capture such patterns, the steady flow branch estimates a nonnegative flow baseline using the static edge representation and the high-order edge representation:
\begin{equation}
\left[
\hat{y}_{ij,t+1}^{s},\ldots,\hat{y}_{ij,t+K}^{s}
\right]
=
\mathrm{MLP}_{bf}
\left(
\tilde{u}_{ij}^{s}
\Vert
u_{ij,t}^{(L)}
\right)
\label{eq:steady_flow_branch}
\end{equation}

Here, $\hat{y}_{ij,t+k}^{s}$ denotes the steady flow component of edge $(v_i,v_j)$ at future time step $t+k$. This branch mainly reflects stable and recurring demand patterns.

\subsubsection{Dynamic Flow Branch}

In addition to stable patterns, OD demand also exhibits short-term fluctuations caused by recent activity changes, intermittent connection states, and sudden demand variations. To capture these dynamics, SAGMTL constructs a task-specific dynamic representation:
\begin{equation}
r_{ij,t}^{d}
=
\mathrm{MLP}_{dyn}
\left(
u_{ij,t}^{d}
\Vert
u_{ij,t}^{(L)}
\right)
\label{eq:dynamic_task_rep}
\end{equation}

Based on this representation, the decoder generates a basic dynamic flow component, a dynamic gate, and a residual variation term:
\begin{equation}
\left[
\tilde{y}_{ij,t+1}^{d},\ldots,\tilde{y}_{ij,t+K}^{d}
\right]
=
\mathrm{MLP}_{d}
\left(
r_{ij,t}^{d}
\right)
\label{eq:basic_dynamic_flow}
\end{equation}

A dynamic gate:
\begin{equation}
\left[
g_{ij,t+1},\ldots,g_{ij,t+K}
\right]
=
\mathrm{MLP}_{g}
\left(
r_{ij,t}^{d}
\Vert
y_{ij,t}
\right)
\label{eq:dynamic_gate}
\end{equation}

And residual term:
\begin{equation}
\left[
\delta_{ij,t+1},\ldots,\delta_{ij,t+K}
\right]
=
\mathrm{MLP}_{\delta}
\left(
r_{ij,t}^{d}
\right)
\label{eq:dynamic_residual}
\end{equation}

The final dynamic flow component is computed as:
\begin{equation}
\hat{y}_{ij,t+k}^{d}
=
\mathrm{softplus}
\left(
\tilde{y}_{ij,t+k}^{d}
+
g_{ij,t+k}
\odot
\delta_{ij,t+k}
\right)
\label{eq:final_dynamic_flow}
\end{equation}

The dynamic gate allows the model to adaptively control the contribution of recent state changes, while the residual term captures short-term flow enhancement, attenuation, and abrupt fluctuations.

\subsubsection{Connection-Guided Flow Fusion}

After obtaining the edge connection probability, steady flow component, and dynamic flow component, SAGMTL uses the predicted connection state to guide the final flow prediction. Specifically, the fusion weight is computed as:
\begin{equation}
\alpha_{ij,t+k}
=
\alpha_0
+
(1-\alpha_0)
\mathrm{MLP}_{\alpha}
\left(
\hat{z}_{ij,t+k}^{e}
\Vert
g_{ij,t+k}
\right)
\label{eq:fusion_weight}
\end{equation}

where $\alpha_0\in[0,1]$ is a predefined lower bound that ensures the dynamic flow branch retains a minimum contribution.

The final OD flow prediction is then given by:
\begin{equation}
\hat{y}_{ij,t+k}
=
(1-\alpha_{ij,t+k})\hat{y}_{ij,t+k}^{s}
+
\alpha_{ij,t+k}\hat{y}_{ij,t+k}^{d}
\label{eq:final_flow_prediction}
\end{equation}

This connection-guided fusion mechanism enables SAGMTL to adaptively balance stable demand patterns and short-term dynamic fluctuations. For OD pairs with regular and high-frequency interactions, the model can rely more on the steady flow component. For intermittent or recently changing OD pairs, the model can increase the contribution of the dynamic flow component, thereby improving prediction robustness under dynamic sparsity.

\subsection{Joint Optimization Objective}

SAGMTL is trained with a joint optimization objective consisting of three complementary parts. The flow prediction loss supervises OD flow intensity estimation, the structural auxiliary loss guides edge-level and node-level activity prediction, and the flow regularization loss improves sparsity awareness, short-term variation modeling, and node-level marginal distribution consistency.

The overall objective is defined as:
\begin{equation}
L
=
L_{\mathrm{flow}}
+
L_{\mathrm{struct}}
+
L_{\mathrm{reg}}
\label{eq:overall_loss}
\end{equation}

where
\begin{IEEEeqnarray}{lCl}
L_{\mathrm{flow}}
&=&
\lambda_{\mathrm{base}}L_{\mathrm{base}}
+
\lambda_{\mathrm{comp}}L_{\mathrm{comp}}
\label{eq:loss_flow}\\
L_{\mathrm{struct}}
&=&
\lambda_{\mathrm{edge}}L_{\mathrm{edge}}
+
\lambda_{\mathrm{node}}L_{\mathrm{node}}
\label{eq:loss_struct}\\
L_{\mathrm{reg}}
&=&
\lambda_{\mathrm{zero}}L_{\mathrm{zero}}
+
\lambda_{\mathrm{vol}}L_{\mathrm{vol}}
+
\lambda_{\mathrm{marg}}L_{\mathrm{marg}}
\label{eq:loss_reg}
\end{IEEEeqnarray}

Here, $\lambda_{\mathrm{base}}$, $\lambda_{\mathrm{comp}}$, $\lambda_{\mathrm{edge}}$, $\lambda_{\mathrm{node}}$, $\lambda_{\mathrm{zero}}$, $\lambda_{\mathrm{vol}}$, and $\lambda_{\mathrm{marg}}$ are weight coefficients used to balance the contributions of different loss terms.

For clarity, we use $\ell_{\mathrm{SL1}}$, $\ell_{\mathrm{BCE}}$, and $\ell_{\mathrm{focal}}$ to denote the Smooth L1 loss, binary cross-entropy (BCE) loss, and focal BCE loss, respectively. Among them, $\ell_{\mathrm{SL1}}$ is used for flow regression and consistency constraints, $\ell_{\mathrm{BCE}}$ is used for node activity prediction, and $\ell_{\mathrm{focal}}$ is used to alleviate the class imbalance problem in edge activity prediction.

For a prediction value $a$ and its ground-truth value $b$, the Smooth L1 loss is defined as\cite{huber1964loss}:
\begin{equation}
\ell_{\mathrm{SL1}}(a,b)
=
\begin{cases}
0.5(a-b)^2, & \text{if } |a-b|<1 \\
|a-b|-0.5, & \text{if } |a-b|\geq 1
\end{cases}
\label{eq:smooth_l1}
\end{equation}

For a predicted probability $p\in[0,1]$ and a binary label $z\in\{0,1\}$, the BCE loss is defined as:
\begin{equation}
\ell_{\mathrm{BCE}}(p,z)
=
-z\log(p)
-
(1-z)\log(1-p)
\label{eq:bce_loss}
\end{equation}

Since edge activity prediction suffers from severe class imbalance, we adopt the focal BCE loss \cite{lin2017focal}:
\begin{equation}
\begin{aligned}
\ell_{\mathrm{focal}}(p,z)
&=
-\alpha_f z(1-p)^{\gamma_f}\log p \\
&\quad
-
(1-\alpha_f)(1-z)p^{\gamma_f}\log(1-p)
\end{aligned}
\label{eq:focal_loss}
\end{equation}

where $\alpha_f$ is the class balancing coefficient and $\gamma_f$ is the focusing parameter. They reduce the loss contribution of easy negative samples and enhance the learning of the minority active-edge samples.

For each future step $t+k$, the active and inactive edge sets are defined as:
\begin{equation}
E_{t+k}^{+}
=
\left\{
(v_i,v_j)\in E
\mid
y_{ij,t+k}>0
\right\}
\label{eq:active_edge_set}
\end{equation}
\begin{equation}
E_{t+k}^{0}
=
\left\{
(v_i,v_j)\in E
\mid
y_{ij,t+k}=0
\right\}
\label{eq:inactive_edge_set}
\end{equation}

\subsubsection{Flow Prediction Loss}

The flow prediction loss supervises the final OD flow prediction. Since OD flow distributions usually exhibit long-tailed patterns, directly computing regression errors in the original flow space may cause high-flow samples to dominate training. Therefore, we adopt the Smooth L1 loss in the logarithmic space:

\begin{equation}
\ell_{ij,t+k}^{f}
=
\ell_{\mathrm{SL1}}
\left(
\log(1+\hat{y}_{ij,t+k}),
\log(1+y_{ij,t+k})
\right)
\label{eq:flow_log_loss}
\end{equation}

where $\hat{y}_{ij,t+k}$ and $y_{ij,t+k}$ denote the predicted and ground-truth OD flows, respectively.

Since this term mainly learns flow intensity when an OD connection actually occurs, the basic flow regression loss is computed over the active edge set:
\begin{equation}
L_{\mathrm{base}}
=
\frac{1}
{\sum_{k=1}^{K}\max\left(|E_{t+k}^{+}|,1\right)}
\sum_{k=1}^{K}
\sum_{i,j:(v_i,v_j)\in E_{t+k}^{+}}
\ell_{ij,t+k}^{f}
\label{eq:base_flow_loss}
\end{equation}

The $\max(\cdot,1)$ operation avoids division by zero when no active edge exists at a certain future step.

\subsubsection{Intensity Compensation Loss}

Although $L_{\mathrm{base}}$ focuses on active OD edges, the model may still tend to fit the majority of low- and medium-flow edges while underestimating high-flow edges. To address this issue, an intensity-aware compensation loss is introduced. For each active OD edge, its relative intensity weight is defined as:

\begin{equation} 
\omega_{ij,t+k} = \frac{ \log\left(1+y_{ij,t+k}\right) }{ \sum_{i',j':(v_{i'},v_{j'})\in E_{t+k}^{+}} \log\left(1+y_{i'j',t+k}\right) } \label{eq:intensity_weight} 
\end{equation}

The intensity compensation loss is defined as:
\begin{equation}
L_{\mathrm{comp}} = \frac{1}{K} \sum_{k=1}^{K} \sum_{i,j:(v_i,v_j)\in E_{t+k}^{+}} \omega_{ij,t+k} \ell_{ij,t+k}^{f} \label{eq:intensity_comp_loss} 
\end{equation}

This term increases the contribution of medium- and high-flow active edges according to their relative flow intensities, thereby alleviating the underestimation of high-intensity OD flows.

\subsubsection{Edge Activity Loss}

The edge activity loss supervises the OD connection state prediction branch. Unlike the flow regression loss, this term is computed over both active and inactive edges, enabling the model to learn a complete binary decision boundary for OD connection activity:
\begin{equation}
L_{\mathrm{edge}} = \frac{1}{K\cdot |E|} \sum_{k=1}^{K} \sum_{i,j:(v_i,v_j)\in E} \ell_{\mathrm{focal}} \left( \hat{z}_{ij,t+k}^{e}, z_{ij,t+k}^{e} \right) \label{eq:edge_activity_loss} 
\end{equation}

Here, $\hat{z}_{ij,t+k}^{e}$ denotes the predicted activation probability of edge $(v_i,v_j)$, and $z_{ij,t+k}^{e}$ denotes the ground-truth edge activity label.

\subsubsection{Node Activity Loss}

The node activity loss provides region-level structural supervision from the perspectives of travel generation and attraction. It measures whether each node is active as an origin or as a destination. It is defined as:
\begin{equation} 
\begin{aligned} 
L_{\mathrm{node}} = \frac{1}{2K|V|} \sum_{k=1}^{K} \sum_{i:v_i\in V} \Big[ & \ell_{\mathrm{BCE}} \left( \hat{z}_{i,t+k}^{n,\mathrm{out}}, z_{i,t+k}^{n,\mathrm{out}} \right) \\ &+ \ell_{\mathrm{BCE}} \left( \hat{z}_{i,t+k}^{n,\mathrm{in}}, z_{i,t+k}^{n,\mathrm{in}} \right) \Big] \end{aligned} \label{eq:node_activity_loss} 
\end{equation}

Here, $\hat{z}_{i,t+k}^{n,out}$ and $\hat{z}_{i,t+k}^{n,in}$ denote the predicted probabilities that node $v_i$ is active as an origin and as a destination, respectively. This auxiliary supervision helps the model capture regional travel generation and attraction patterns, which further benefits sparse OD connection identification.

\subsubsection{Zero-Edge Suppression Loss}

In sparse OD systems, most OD edges remain inactive at many time steps. To reduce false positive flow predictions on zero-flow edges, we introduce a zero-edge suppression loss:
\begin{equation}
\begin{aligned}
L_{\mathrm{zero}}
&=
\frac{1}
{\sum_{k=1}^{K}\max\left(|E_{t+k}^{0}|,1\right)}
\\
&\quad \times
\sum_{k=1}^{K}
\sum_{i,j:(v_i,v_j)\in E_{t+k}^{0}}
\ell_{\mathrm{SL1}}
\left(
\hat{y}_{ij,t+k},0
\right).
\end{aligned}
\label{eq:zero_edge_loss}
\end{equation}

This term penalizes nonzero predictions on truly inactive OD edges and improves the model’s sparsity awareness.

\subsubsection{Volatility Consistency Loss}

To improve the model's ability to capture short-term flow changes, we  introduce a volatility consistency loss. The ground-truth flow variation is defined as:
\begin{equation}
\Delta y_{ij,t+k}
=
y_{ij,t+k}
-
y_{ij,t+k-1}
\label{eq:true_flow_variation}
\end{equation}

The predicted flow variation is defined as:
\begin{equation}
\Delta \hat{y}_{ij,t+k}
=
\hat{y}_{ij,t+k}
-
\hat{y}_{ij,t+k-1}
\label{eq:pred_flow_variation}
\end{equation}

that is, the currently observed ground-truth flow is used as the starting point for computing the predicted variation.

The set of edges with significant flow variations is defined as:
\begin{equation}
E_{t+k}^{\mathrm{var}}
=
\left\{
(v_i,v_j)\in E
\mid
\left|\Delta y_{ij,t+k}\right|>\varepsilon
\right\}
\label{eq:variation_edge_set}
\end{equation}

where $\varepsilon$ is a variation threshold. The volatility consistency loss is defined as:

\begin{equation}
\begin{aligned}
L_{\mathrm{vol}}
&=
\frac{1}
{\sum_{k=1}^{K}\max\left(|E_{t+k}^{\mathrm{var}}|,1\right)}
\\
&\quad \times
\sum_{k=1}^{K}
\sum_{i,j:(v_i,v_j)\in E_{t+k}^{\mathrm{var}}}
\ell_{\mathrm{SL1}}
\left(
\Delta \hat{y}_{ij,t+k},
\Delta y_{ij,t+k}
\right)
\end{aligned}
\label{eq:volatility_loss}
\end{equation}

This term encourages the model to capture both the direction and magnitude of short-term OD flow changes, especially for OD pairs with significant temporal fluctuations.

\subsubsection{Node Inflow--Outflow Marginal Consistency Loss}

OD prediction should preserve not only edge-level flow accuracy but also node-level distribution consistency. To reduce systematic bias in regional total inflows and outflows, we introduce a node inflow–outflow marginal consistency loss.

For any node $v_i$, the predicted and ground-truth total outflow at time step $t+k$ are defined as:
\begin{align}
\hat{o}_{i,t+k}
&=
\sum_{j:(v_i,v_j)\in E}
\hat{y}_{ij,t+k} \\
o_{i,t+k}
&=
\sum_{j:(v_i,v_j)\in E}
y_{ij,t+k}
\label{eq:node_outflow}
\end{align}

Similarly the predicted and ground-truth total inflow are defined as:
\begin{align}
\hat{q}_{i,t+k}
&=
\sum_{j:(v_j,v_i)\in E}
\hat{y}_{ji,t+k} \\
q_{i,t+k}
&=
\sum_{j:(v_j,v_i)\in E}
y_{ji,t+k}
\label{eq:node_inflow}
\end{align}

The node inflow--outflow marginal consistency loss is defined as:
\begin{equation}
\begin{aligned}
L_{\mathrm{marg}}
=
\frac{1}{2K|V|}
\sum_{k=1}^{K}
\sum_{i:v_i\in V}
\Big[
&
\ell_{\mathrm{SL1}}
\left(
\hat{o}_{i,t+k},
o_{i,t+k}
\right)
\\
&+
\ell_{\mathrm{SL1}}
\left(
\hat{q}_{i,t+k},
q_{i,t+k}
\right)
\Big]
\end{aligned}
\label{eq:consistency_loss}
\end{equation}

This constraint encourages the model to maintain consistency between predicted and ground-truth regional travel generation and attraction, thereby improving the stability of the overall OD flow distribution.

In summary, the proposed joint optimization objective combines the main flow prediction task, structural auxiliary tasks, and sparsity-aware flow regularization within a unified training framework. By jointly optimizing these complementary objectives, SAGMTL can better adapt to long-tailed flow distributions, intermittent OD activations, and node-level marginal flow imbalance in dynamic sparse OD demand prediction.

\section{EXPERIMENTS}
\subsection{Datasets}

We evaluate SAGMTL on three real-world urban mobility datasets: Beijing Taxi, Nanjing Taxi, and Chengdu Ride-hailing. The Beijing dataset contains taxi GPS trajectories from November 1 to November 30, 2012; the Nanjing dataset contains taxi GPS trajectories from January 13 to February 2, 2011; and the Chengdu dataset contains ride-hailing trajectories from November 1 to November 30, 2016. For each city, we determine the study area based on the spatial coverage of valid trips and divide it into regular 1.5 km $\times$ 1.5 km grids. Each grid cell is treated as a traffic analysis zone and can serve as either an origin or a destination.

We map each trip origin and destination to the corresponding grid cells and count directed trips between grid pairs at 20-minute intervals to construct dynamic OD flow sequences. Records with missing coordinates, abnormal timestamps, or O/D outside the study area are removed. The resulting directed trip counts are used as the ground truth for OD flow prediction. To avoid test-set information leakage, the OD edge set $E$ is constructed from directed region pairs with valid trip records in the training set and is kept fixed for validation and testing.

To capture regional functional semantics, we collect POI data through the AMAP API and map POIs to grid cells by location. For each region, POI counts across categories such as catering, shopping, accommodation, healthcare, transportation, public facilities, enterprises, life services, commercial residences, sports and leisure, government agencies, and scenic spots are used to form a multidimensional functional feature vector.

Table~\ref{tab:dataset_statistics} reports the statistics of the three datasets, including the numbers of nodes and historical unique OD edges, the average number of active edges per time slot, the active-edge density, and the average total flow. The active-edge density is defined as the average proportion of active OD edges among all possible directed non-self OD pairs in each time slot. As shown in Table~\ref{tab:dataset_statistics}, the active-edge density is below 0.6\% for all datasets, indicating that most OD pairs remain inactive at any given time and that OD demand exhibits strong dynamic sparsity.

\begin{table}[!t]
\caption{Statistics of the dynamic sparse OD datasets.}
\label{tab:dataset_statistics}
\centering
\scriptsize
\setlength{\tabcolsep}{2.6pt}
\renewcommand{\arraystretch}{1.08}
\resizebox{\columnwidth}{!}{
\begin{tabular}{lccccc}
\hline
Dataset & Nodes & Unique E & Avg. E/S & Density & Avg. Flow \\
\hline
Beijing & 675 & 159765 & 2300 & 0.0051 & 3064 \\
Chengdu & 673 & 50703  & 1754 & 0.0039 & 2690 \\
Nanjing & 673 & 52603  & 1624 & 0.0036 & 3144 \\
\hline
\end{tabular}
}
\end{table}

\subsection{Baselines}

To comprehensively evaluate the effectiveness of the proposed method, we compare SAGMTL with representative baselines from four categories:
\begin{enumerate}
\item Traditional statistical and machine learning methods: HA (Historical Average) and SVR (Support Vector Regression).
\item Deep temporal models: GRU \cite{cho2014gru} and LSTM.
\item Spatiotemporal graph forecasting models: STGCN \cite{yu2018stgcn}, DCRNN \cite{li2018dcrnn}, and AGCRN \cite{bai2020agcrn}.
\item OD prediction and sparse demand modeling methods: CSTN \cite{liu2019cstn}, GEML \cite{wang2019geml}, MPGCN \cite{shi2020mpgcn}, TGAE \cite{wang2023tgae}, STZINB-GNN \cite{zhuang2022uncertainty}, and DSTCN \cite{gong2026dstcn}.
\end{enumerate}

For a fair comparison, all methods use the same training, validation, and test splits and are trained and evaluated on the same OD flow sequences. For spatiotemporal graph forecasting models originally designed for node-level traffic prediction, we concatenate the learned origin-node embedding and destination-node embedding to construct an OD-pair feature representation. Each OD pair is then treated as a prediction unit, and the models are trained and tested under the same regional partition and OD relation graph. For HA, SVR, and GRU, modeling is conducted directly on the same OD-pair flow sequences.

\subsection{Evaluation Metrics}

We evaluate the model from two perspectives: OD flow prediction accuracy and structural state identification ability.

\subsubsection{Flow Prediction Metrics}

To evaluate OD flow prediction accuracy, we adopt Mean Absolute Error (MAE), Root Mean Square Error (RMSE), and Mean Absolute Percentage Error (MAPE). Since OD flow matrices contain a large number of zero values, MAE, RMSE, and MAPE are computed on the set of truly active OD edges.

Let $T_{\mathrm{test}}$ denote the set of prediction starting time steps in the test set, and the total number of active-edge samples during testing is defined as
\begin{equation}
N_{\mathrm{test}}^{+}
=
\sum_{t\in T_{\mathrm{test}}}
\sum_{k=1}^{K}
\left|E_{t+k}^{+}\right|
\label{eq:test_active_samples}
\end{equation}

MAE is defined as:
\begin{equation}
\mathrm{MAE}
=
\frac{1}{N_{\mathrm{test}}^{+}}
\sum_{t\in T_{\mathrm{test}}}
\sum_{k=1}^{K}
\sum_{(v_i,v_j)\in E_{t+k}^{+}}
\left|
\hat{y}_{ij,t+k}
-
y_{ij,t+k}
\right|
\label{eq:mae}
\end{equation}

RMSE is defined as:
\begin{equation}
\begin{aligned}
\mathrm{RMSE}
&=
\Bigg(
\frac{1}{N_{\mathrm{test}}^{+}}
\sum_{t\in T_{\mathrm{test}}}
\sum_{k=1}^{K}
\\
&\quad
\sum_{\substack{(v_i,v_j)\in E_{t+k}^{+}}}
\left(
\hat{y}_{ij,t+k}
-
y_{ij,t+k}
\right)^2
\Bigg)^{\frac{1}{2}} 
\end{aligned}
\label{eq:rmse}
\end{equation}

MAPE is defined as:
\begin{equation}
\begin{aligned}
\mathrm{MAPE}
&=
\frac{100\%}{N_{\mathrm{test}}^{+}}
\sum_{t\in T_{\mathrm{test}}}
\sum_{k=1}^{K}
\\
&\quad
\sum_{\substack{(v_i,v_j)\in E_{t+k}^{+}}}
\frac{
\left|
\hat{y}_{ij,t+k}
-
y_{ij,t+k}
\right|
}{
\left|y_{ij,t+k}\right|
}
\end{aligned}
\label{eq:mape}
\end{equation}

\subsubsection{Structural Prediction Metrics}

For edge activity identification, we use F1-score as the evaluation metric, denoted as $\mathrm{F1}_{\mathrm{edge}}$. For node activity prediction, we report the average F1-score over origin-side and destination-side activity prediction:
\begin{equation}
\mathrm{F1}_{\mathrm{node}}
=
\frac{1}{2}
\left(
\mathrm{F1}_{\mathrm{node}}^{\mathrm{out}}
+
\mathrm{F1}_{\mathrm{node}}^{\mathrm{in}}
\right)
\label{eq:node_f1}
\end{equation}

where $\mathrm{F1}_{\mathrm{node}}^{\mathrm{out}}$ and $\mathrm{F1}_{\mathrm{node}}^{\mathrm{in}}$ denote the F1-scores for node activity prediction as origins and destinations, respectively.

\subsection{Experimental Settings}

All models are implemented in PyTorch and trained and tested on a server equipped with an NVIDIA GeForce RTX 3090 GPU. Each dataset is split into training, validation, and test sets in chronological order with a ratio of 7:1:2. For a fair comparison, all temporal models use the same input window length $T=18$ and prediction horizon $K=6$.

We adopt a unified data preprocessing pipeline and construct the OD edge set in the same way for all methods. All methods are evaluated on the same test set. For spatiotemporal graph models, the input graphs are constructed based on the same regional partition and OD relations. For GRU, SVR, and HA, training and testing are conducted on the same OD-pair flow sequences.

The loss weights of SAGMTL are set as follows: $\lambda_{\mathrm{base}}=3.0$ for the basic flow prediction loss, $\lambda_{\mathrm{comp}}=1.0$ for the intensity compensation loss, $\lambda_{\mathrm{edge}}=0.08$ for the edge activity supervision loss, $\lambda_{\mathrm{node}}=0.08$ for the node activity supervision loss, $\lambda_{\mathrm{zero}}=0.035$ for the zero-edge suppression loss, $\lambda_{\mathrm{vol}}=0.1$ for the volatility consistency loss, and $\lambda_{\mathrm{marg}}=0.1$ for the marginal consistency loss.

\subsection{Overall Prediction Performance Comparison}

Table~\ref{tab:overall_performance} reports the OD demand prediction results on the Beijing, Chengdu, and Nanjing datasets. Overall, SAGMTL achieves the best MAE and MAPE on all three datasets, demonstrating its effectiveness in modeling the spatiotemporal evolution of dynamic sparse OD demand.

\begin{table*}[!t]
\caption{Overall prediction performance comparison on three city datasets.}
\label{tab:overall_performance}
\centering
\footnotesize
\renewcommand{\arraystretch}{1.05}
\begin{tabular*}{\textwidth}{@{\extracolsep{\fill}}lccccccccc@{}}
\hline
\multirow{2}{*}{Model}
& \multicolumn{3}{c}{Beijing}
& \multicolumn{3}{c}{Chengdu}
& \multicolumn{3}{c}{Nanjing} \\
\cline{2-10}
& MAE & RMSE & MAPE
& MAE & RMSE & MAPE
& MAE & RMSE & MAPE \\
\hline
HA         & 1.1682 & 1.6905 & 50.6618 & 1.1370 & 1.5809 & 47.0994 & 1.9119 & 4.4453 & 56.0082 \\
SVR        & 1.1217 & 1.6935 & 51.2063 & 1.1304 & 1.5097 & 46.9262 & 1.9294 & 4.8357 & 57.7905 \\
GRU        & 1.1191 & 1.6066 & 51.3054 & 1.0901 & 1.3990 & 46.5434 & 1.8011 & 4.3153 & 50.2860 \\
LSTM       & 1.1112 & 1.6018 & 50.8970 & 1.0965 & 1.4037 & 46.7260 & 1.8446 & 4.2964 & 52.4634 \\
STGCN      & 1.1261 & 1.6130 & 51.6058 & 1.0756 & 1.3859 & 45.4418 & 1.8220 & 4.1084 & 51.1985 \\
DCRNN      & 1.0887 & 1.5853 & 49.8150 & 1.0075 & 1.3017 & 42.8198 & 1.7316 & 4.0866 & 48.6876 \\
AGCRN      & 1.0470 & 1.5436 & 51.2921 & 1.0911 & 1.4208 & 45.8805 & 1.8674 & 4.2297 & 52.4742 \\
CSTN       & 1.0597 & 1.5624 & 48.4493 & 1.0061 & 1.2822 & 43.1267 & 1.7534 & 4.0705 & 49.5929 \\
GEML       & 1.0437 & 1.5050 & 48.0701 & 1.0031 & 1.2521 & 43.0421 & 1.6771 & 4.2529 & 47.8914 \\
MPGCN      & 1.0351 & 1.4974 & 47.6747 & 0.9919 & 1.2286 & 43.0144 & 1.6744 & 4.2413 & 47.4482 \\
TGAE       & 1.0380 & 1.4967 & 47.8686 & 0.9838 & 1.2193 & 42.7262 & 1.6882 & 4.2793 & 48.0926 \\
STZINB-GNN & 0.9857 & 1.5129 & 45.0495 & 0.9336 & 1.1952 & 40.2827 & 1.7188 & \textbf{4.0254} & 48.4879 \\
DSTCN      & 1.0877 & 1.5859 & 49.7298 & 1.0399 & 1.3405 & 44.0502 & 1.7158 & 4.0534 & 48.1089 \\
SAGMTL     & \textbf{0.8692} & \textbf{1.4505} & \textbf{39.2874}
           & \textbf{0.8892} & \textbf{1.1777} & \textbf{37.7583}
           & \textbf{1.6135} & 4.2491 & \textbf{41.6601} \\
\hline
\end{tabular*}
\end{table*}

On the Beijing dataset, SAGMTL achieves an MAE of 0.8692, reducing the error by 11.82\% compared with the best baseline STZINB-GNN (0.9857). Its MAPE decreases from 45.05\% to 39.29\%, with a relative reduction of 12.79\%. On the Chengdu dataset, SAGMTL achieves an MAE of 0.8892 and a MAPE of 37.76\%, improving over STZINB-GNN by 4.76\% and 6.27\%, respectively. On the Nanjing dataset, SAGMTL obtains the lowest MAE (1.6135) and MAPE (41.66\%), reducing the errors by 3.64\% and 12.20\% compared with the best baseline MPGCN.

Although SAGMTL has a slightly higher RMSE than some baselines on the Nanjing dataset, it achieves the best MAE and MAPE. This indicates that SAGMTL predicts most OD flows more accurately, while RMSE is more sensitive to a small number of high-flow outliers. Therefore, SAGMTL better balances overall prediction accuracy and long-tailed flow modeling.

Compared with traditional methods such as HA and SVR, SAGMTL better captures nonlinear spatiotemporal dependencies. Compared with GRU, it further incorporates spatial and structural information. Although graph-based models such as STGCN, DCRNN, and AGCRN can model spatial dependencies, they are originally designed for node-level traffic prediction and are less effective in representing edge-level OD interactions. OD-specific models such as GEML, MPGCN, TGAE, and STZINB-GNN achieve stronger performance by modeling inter-regional relationships or sparse zero-flow patterns. However, most of them still treat OD prediction as a single flow regression problem.

In contrast, SAGMTL explicitly decomposes OD demand prediction into connection state identification and flow intensity estimation. By introducing edge and node activity prediction as auxiliary tasks and using node--edge collaborative updates, SAGMTL learns more robust structural representations, leading to superior prediction performance across the three datasets.

\subsection{Multi-step Prediction Performance Analysis}

Tables~\ref{tab:multistep_beijing}--\ref{tab:multistep_nanjing} report the MAE results of different models over the next six prediction steps. Overall, most baselines show increasing errors as the prediction horizon becomes longer, indicating more severe error accumulation in long-term forecasting. For example, on the Beijing dataset, the MAE of HA increases from 1.1557 to 1.1803; on the Nanjing dataset, the MAE of GRU increases from 1.7171 to 1.8624, with a relative increase of 8.46\%.

\begin{table}[!t]
\caption{Multi-step MAE comparison on the Beijing dataset.}
\label{tab:multistep_beijing}
\centering
\scriptsize
\setlength{\tabcolsep}{3.2pt}
\renewcommand{\arraystretch}{1.08}
\resizebox{\columnwidth}{!}{
\begin{tabular}{lcccccc}
\hline
Model & H1 & H2 & H3 & H4 & H5 & H6 \\
\hline
HA         & 1.1557 & 1.1612 & 1.1660 & 1.1705 & 1.1753 & 1.1803 \\
SVR        & 1.1090 & 1.1147 & 1.1177 & 1.1247 & 1.1306 & 1.1333 \\
GRU        & 1.1085 & 1.1127 & 1.1188 & 1.1217 & 1.1247 & 1.1285 \\
LSTM       & 1.0963 & 1.1036 & 1.1091 & 1.1132 & 1.1202 & 1.1249 \\
STGCN      & 1.1137 & 1.1203 & 1.1234 & 1.1275 & 1.1332 & 1.1384 \\
DCRNN      & 1.0824 & 1.0861 & 1.0880 & 1.0899 & 1.0922 & 1.0936 \\
AGCRN      & 1.0418 & 1.0448 & 1.0456 & 1.0466 & 1.0510 & 1.0521 \\
CSTN       & 1.0636 & 1.0613 & 1.0616 & 1.0595 & 1.0576 & 1.0545 \\
GEML       & 1.0428 & 1.0415 & 1.0396 & 1.0413 & 1.0471 & 1.0498 \\
MPGCN      & 1.0259 & 1.0290 & 1.0403 & 1.0375 & 1.0377 & 1.0404 \\
TGAE       & 1.0351 & 1.0359 & 1.0354 & 1.0369 & 1.0417 & 1.0429 \\
STZINB-GNN & 0.9758 & 0.9792 & 0.9817 & 0.9865 & 0.9928 & 0.9984 \\
DSTCN      & 1.0878 & 1.0866 & 1.0904 & 1.0866 & 1.0855 & 1.0894 \\
SAGMTL     & \textbf{0.8706} & \textbf{0.8694} & \textbf{0.8695} & \textbf{0.8695} & \textbf{0.8694} & \textbf{0.8665} \\
\hline
\end{tabular}
}
\end{table}

\begin{table}[!t]
\caption{Multi-step MAE comparison on the Chengdu dataset.}
\label{tab:multistep_chengdu}
\centering
\scriptsize
\setlength{\tabcolsep}{3.2pt}
\renewcommand{\arraystretch}{1.08}
\resizebox{\columnwidth}{!}{
\begin{tabular}{lcccccc}
\hline
Model & H1 & H2 & H3 & H4 & H5 & H6 \\
\hline
HA         & 1.1021 & 1.1163 & 1.1304 & 1.1443 & 1.1578 & 1.1709 \\
SVR        & 1.0966 & 1.1130 & 1.1236 & 1.1385 & 1.1503 & 1.1607 \\
GRU        & 1.0526 & 1.0674 & 1.0855 & 1.0978 & 1.1115 & 1.1260 \\
LSTM       & 1.0565 & 1.0728 & 1.0854 & 1.1019 & 1.1230 & 1.1398 \\
STGCN      & 1.0664 & 1.0664 & 1.0753 & 1.0747 & 1.0825 & 1.0884 \\
DCRNN      & 1.0145 & 1.0086 & 1.0064 & 1.0044 & 1.0035 & 1.0077 \\
AGCRN      & 1.0716 & 1.0812 & 1.0910 & 1.0972 & 1.1008 & 1.1049 \\
CSTN       & 1.0159 & 1.0069 & 1.0008 & 1.0016 & 1.0042 & 1.0070 \\
GEML       & 1.0039 & 0.9983 & 0.9978 & 0.9973 & 0.9989 & 1.0070 \\
MPGCN      & 0.9824 & 0.9852 & 0.9883 & 0.9930 & 0.9984 & 1.0040 \\
TGAE       & 0.9791 & 0.9823 & 0.9831 & 0.9846 & 0.9857 & 0.9882 \\
STZINB-GNN & 0.9277 & 0.9333 & 0.9346 & 0.9370 & 0.9355 & 0.9336 \\
DSTCN      & 1.0207 & 1.0260 & 1.0276 & 1.0426 & 1.0572 & 1.0654 \\
SAGMTL     & \textbf{0.8875} & \textbf{0.8927} & \textbf{0.8918} & \textbf{0.8909} & \textbf{0.8901} & \textbf{0.8821} \\
\hline
\end{tabular}
}
\end{table}

\begin{table}[!t]
\caption{Multi-step MAE comparison on the Nanjing dataset.}
\label{tab:multistep_nanjing}
\centering
\scriptsize
\setlength{\tabcolsep}{3.2pt}
\renewcommand{\arraystretch}{1.08}
\resizebox{\columnwidth}{!}{
\begin{tabular}{lcccccc}
\hline
Model & H1 & H2 & H3 & H4 & H5 & H6 \\
\hline
HA         & 1.8195 & 1.8635 & 1.9000 & 1.9329 & 1.9631 & 1.9924 \\
SVR        & 1.8337 & 1.8789 & 1.9174 & 1.9466 & 1.9895 & 2.0105 \\
GRU        & 1.7171 & 1.7647 & 1.7944 & 1.8256 & 1.8427 & 1.8624 \\
LSTM       & 1.7415 & 1.7706 & 1.7988 & 1.8340 & 1.8578 & 1.9262 \\
STGCN      & 1.7286 & 1.7706 & 1.7988 & 1.8340 & 1.8578 & 1.8764 \\
DCRNN      & 1.6531 & 1.7020 & 1.7281 & 1.7434 & 1.7676 & 1.7955 \\
AGCRN      & 1.7859 & 1.8330 & 1.8607 & 1.8742 & 1.9044 & 1.9470 \\
CSTN       & 1.6794 & 1.7175 & 1.7461 & 1.7665 & 1.7936 & 1.8177 \\
GEML       & 1.6475 & 1.6578 & 1.6742 & 1.6841 & 1.6957 & 1.7035 \\
MPGCN      & 1.6333 & 1.6569 & 1.6743 & 1.6830 & 1.6968 & 1.7024 \\
TGAE       & 1.6394 & 1.6682 & 1.6847 & 1.7029 & 1.7154 & 1.7189 \\
STZINB-GNN & 1.6183 & 1.6764 & 1.7112 & 1.7458 & 1.7776 & 1.7838 \\
DSTCN      & 1.6243 & 1.6697 & 1.7009 & 1.7331 & 1.7611 & 1.8062 \\
SAGMTL     & \textbf{1.5698} & \textbf{1.5959} & \textbf{1.6120} & \textbf{1.6265} & \textbf{1.6364} & \textbf{1.6409} \\
\hline
\end{tabular}
}
\end{table}

In contrast, SAGMTL maintains more stable performance across all three datasets. On the Beijing dataset, its MAE remains within 0.8665--0.8706, with a fluctuation range of only 0.0041. On the Chengdu dataset, the MAE stays around 0.88, with an overall variation below 0.011. Although the error also increases on the Nanjing dataset, its growth is smaller than that of most baselines.

At the sixth prediction step (H6), SAGMTL still achieves clear advantages. It obtains an MAE of 0.8665 on the Beijing dataset, reducing the error by 13.21\% compared with STZINB-GNN (0.9984). On the Chengdu and Nanjing datasets, SAGMTL reduces MAE by 5.52\% and 3.61\% compared with the best baselines, respectively.

These results demonstrate that SAGMTL has stronger long-term prediction stability. The node--edge collaborative update mechanism captures the coupling between regional states and OD interactions, while edge and node activity supervision provides structural priors that help reduce error propagation in multi-step forecasting.

\subsection{Visualization of Representative OD Pairs}

\begin{figure*}[!t]
\centering
\includegraphics[width=0.98\textwidth]{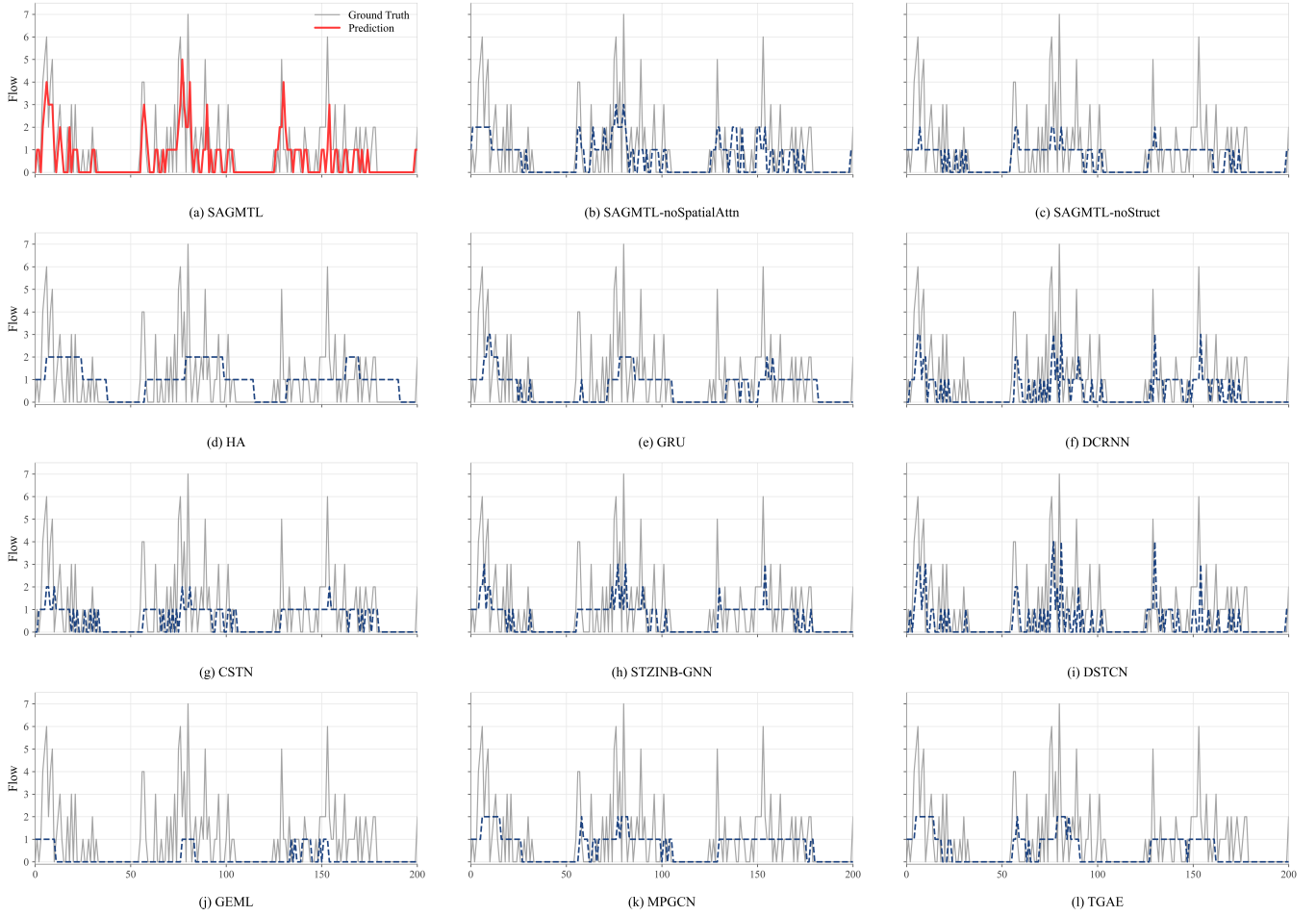}
\caption{Fitting results on a representative stable OD edge. 
The gray curve denotes the ground-truth flow, the red curve denotes the prediction of SAGMTL, and the blue dashed curves denote the predictions of comparison methods or ablated variants.}
\label{fig:balanced_edge}
\end{figure*}

\begin{figure*}[!t]
\centering
\includegraphics[width=0.98\textwidth]{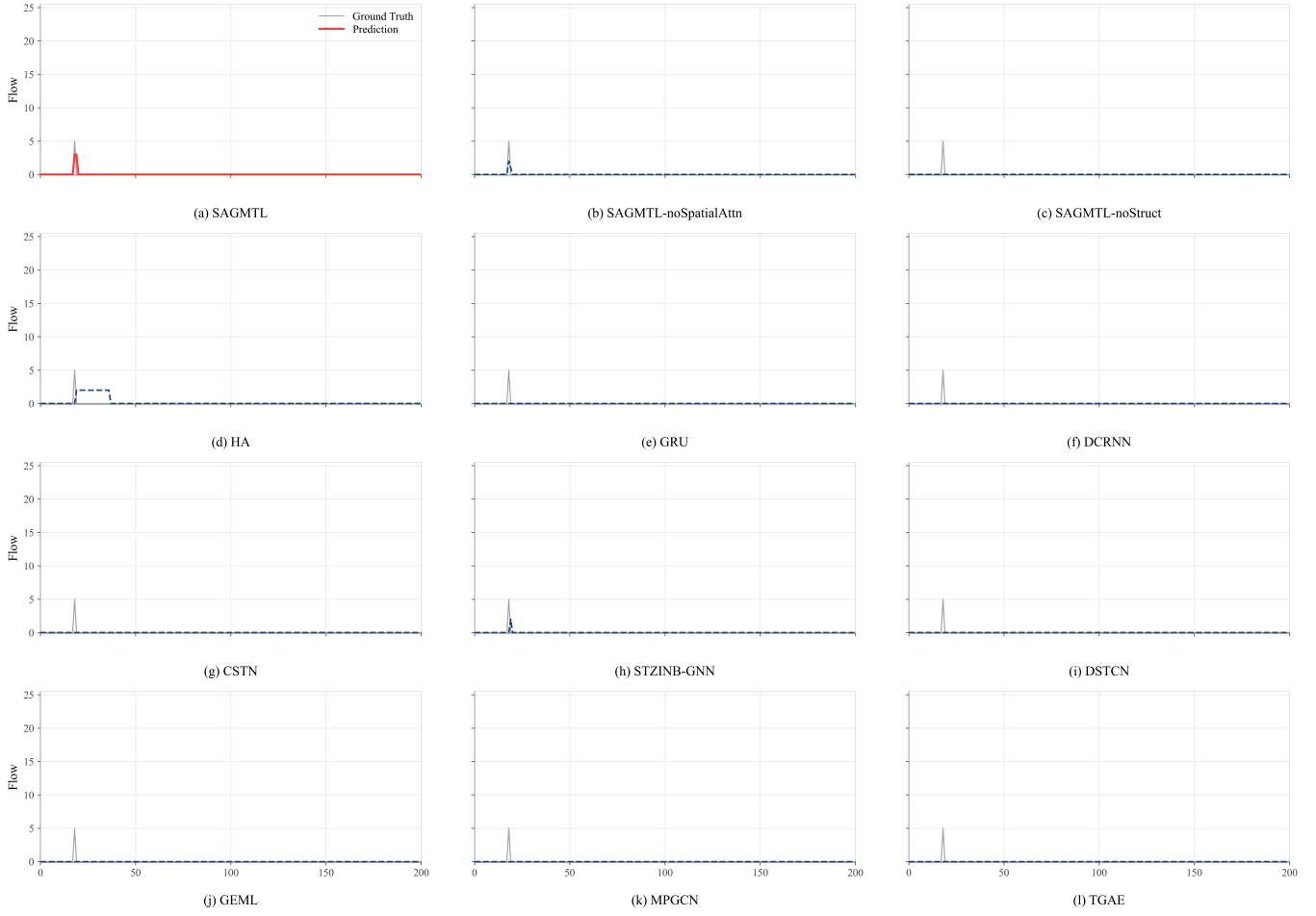}
\caption{Fitting results on a representative bursty OD edge. 
The gray curve denotes the ground-truth flow, the red curve denotes the prediction of SAGMTL, and the blue dashed curves denote the predictions of comparison methods or ablated variants.}
\label{fig:burst_edge}
\end{figure*}

\begin{figure*}[!t]
\centering
\includegraphics[width=0.98\textwidth]{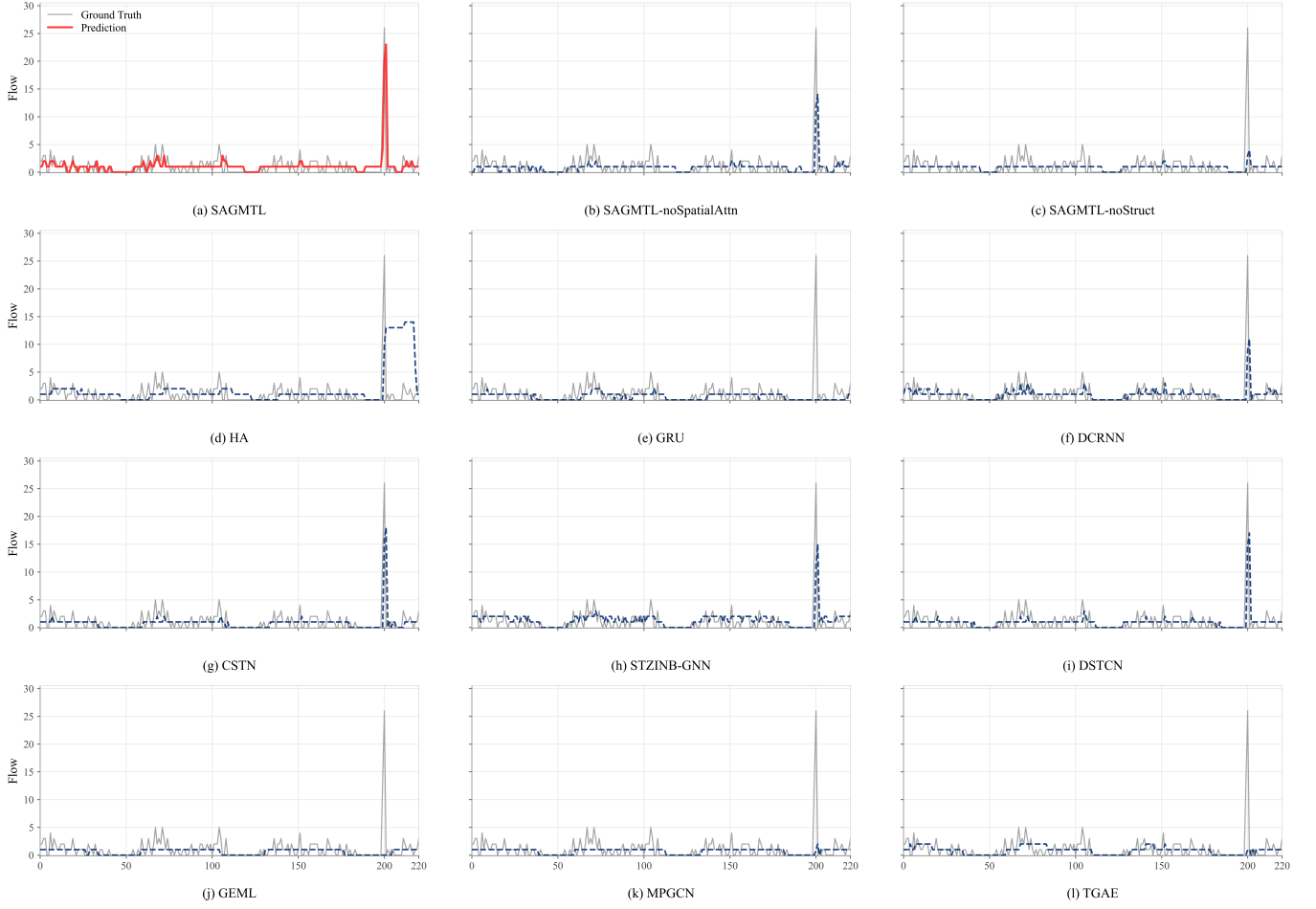}
\caption{Fitting results on a representative highly volatile OD edge. 
The gray curve denotes the ground-truth flow, the red curve denotes the prediction of SAGMTL, and the blue dashed curves denote the predictions of comparison methods or ablated variants.}
\label{fig:volatile_edge}
\end{figure*}

To further examine how different models capture dynamic sparse OD patterns, we select three representative OD pairs from the Beijing test set, corresponding to balanced, bursty, and highly volatile travel patterns. Figs.~\ref{fig:balanced_edge}--\ref{fig:volatile_edge} show the ground-truth flow series and the predictions of different models. In these figures, the gray curves denote the ground-truth flows, the red curves denote the predictions of SAGMTL, and the blue dashed curves denote the predictions of comparison methods or ablated variants.

For the stable OD pair in Fig.~\ref{fig:balanced_edge}, most methods capture the overall trend, but traditional statistical and temporal models tend to over-smooth local fluctuations, while some spatiotemporal graph models still generate false flows during inactive periods. In contrast, SAGMTL better distinguishes active and inactive states and suppresses false positives in zero-flow intervals.

For the bursty OD pair in Fig.~\ref{fig:burst_edge}, most baselines underestimate sudden flow increases or fail to capture peak timing. SAGMTL better identifies the connection reactivation process and achieves more accurate predictions around peak periods, showing stronger sensitivity to low-frequency and bursty OD interactions.

For the highly volatile OD pair in Fig.~\ref{fig:volatile_edge}, baseline methods generally capture the overall direction but show lag or amplitude attenuation when modeling continuous peaks and troughs. SAGMTL tracks the dynamic variations more accurately and maintains higher consistency with the ground-truth sequence.

Overall, these visualization results show that SAGMTL can model stable trends, connection activation, and abrupt flow changes more effectively. This further validates the benefit of jointly modeling connection states and flow intensities in dynamic sparse OD demand prediction.

\subsection{Ablation Study}

To evaluate the contribution of each component in SAGMTL, we conduct ablation experiments on the Beijing dataset, as reported in Table~\ref{tab:ablation_study}. We remove the structural auxiliary tasks, spatial distance attention module, dynamic gating mechanism, node--edge collaborative update module, and steady--dynamic flow decomposition decoder, respectively, and compare their effects on prediction performance.

\begin{table}[!t]
\caption{Ablation study on the Beijing dataset.}
\label{tab:ablation_study}
\centering
\scriptsize
\setlength{\tabcolsep}{3.0pt}
\renewcommand{\arraystretch}{1.08}
\resizebox{\columnwidth}{!}{
\begin{tabular}{lccccc}
\hline
Model & MAE & RMSE & MAPE & Node F1 & Edge F1 \\
\hline
SAGMTL-noStruct      & 0.9105 & 1.4709 & 41.3625 & 0.5260 & 0.0342 \\
SAGMTL-noSpatialAttn & 0.8876 & 1.4608 & 40.2677 & 0.8462 & 0.3327 \\
SAGMTL-noDynGate     & 0.9098 & 1.4728 & 41.3015 & 0.8491 & 0.3358 \\
SAGMTL-noCO          & 0.8909 & 1.4616 & 40.3906 & 0.8421 & 0.3291 \\
SAGMTL-noSF          & 0.9146 & 1.4796 & 41.4413 & 0.8484 & 0.3327 \\
SAGMTL-singleHead    & 0.8909 & 1.4616 & 40.3906 & \textbf{0.8521} & 0.3291 \\
SAGMTL-noEdgeAct     & 0.8794 & 1.4573 & 39.7602 & 0.8453 & 0.0350 \\
SAGMTL-noNodeAct     & 0.8729 & 1.4557 & 39.4891 & 0.5703 & 0.3289 \\
SAGMTL               & \textbf{0.8692} & \textbf{1.4505} & \textbf{39.2874} & 0.8514 & \textbf{0.3383} \\
\hline
\end{tabular}
}
\end{table}

\textbf{Effect of Structural Auxiliary Tasks:} Removing the edge activity prediction branch (SAGMTL-noEdgeAct) increases MAE from 0.8692 to 0.8794 and MAPE from 39.29\% to 39.76\%, showing that edge activity supervision helps identify potentially active OD connections. Removing the node activity prediction branch (SAGMTL-noNodeAct) increases MAE to 0.8729, indicating that node-level activity information also contributes to regional state modeling.When both auxiliary tasks are removed (SAGMTL-noStruct), MAE rises to 0.9105 and MAPE increases by 2.08 percentage points, demonstrating that structural supervision provides important priors for dynamic sparse OD prediction.

\textbf{Effect of the Spatial Distance Attention:} SAGMTL-noSpatialAttn removes the spatial distance attention module. Its MAE increases from 0.8692 to 0.8876, and MAPE increases from 39.29\% to 40.27\%. This confirms that geographical proximity remains important for modeling inter-regional travel demand, and spatial distance attention helps capture more reasonable spatial dependencies.

\textbf{Effect of the Node--Edge Collaborative Update:} SAGMTL-noCO removes the node--edge collaborative update mechanism. The MAE increases to 0.8909, and MAPE increases to 40.39\%. This suggests that independent node and edge encoding is insufficient, while node--edge collaborative updates better capture the coupling between regional states and OD interactions.

\textbf{Effect of the Dynamic Gating Mechanism:} SAGMTL-noDynGate removes the gating mechanism in the dynamic flow branch. The MAE increases to 0.9098, and MAPE increases to 41.30\%, showing a substantial performance drop. This indicates that dynamic gating is crucial for adaptively modulating flow prediction according to connection states.

\textbf{Effect of the Steady--Dynamic Flow Decoder Structure:} To evaluate the effectiveness of the steady--dynamic flow decomposition decoder, we construct two variants: SAGMTL-noSF and SAGMTL-singleHead. SAGMTL-noSF removes the steady-flow branch and only keeps the dynamic-flow branch. SAGMTL-singleHead keeps the structural auxiliary branches unchanged but replaces the steady--dynamic flow decomposition decoder with a single MLP prediction head, which directly maps the $L$-th layer edge representation to the final OD flow prediction. The results show that both variants lead to clear performance degradation. SAGMTL-noSF obtains the worst MAE of 0.9146, while SAGMTL-singleHead increases the MAE to 0.8909. These results indicate that explicitly decomposing OD demand into steady and dynamic components is more effective than directly predicting flows from edge representations with a single head. The steady--dynamic flow decomposition decoder helps the model jointly capture relatively stable demand patterns and short-term flow fluctuations, thereby improving OD flow prediction accuracy.

\subsection{Parameter Sensitivity Analysis}

\begin{figure*}[!t]
\centering
\includegraphics[width=0.98\textwidth]{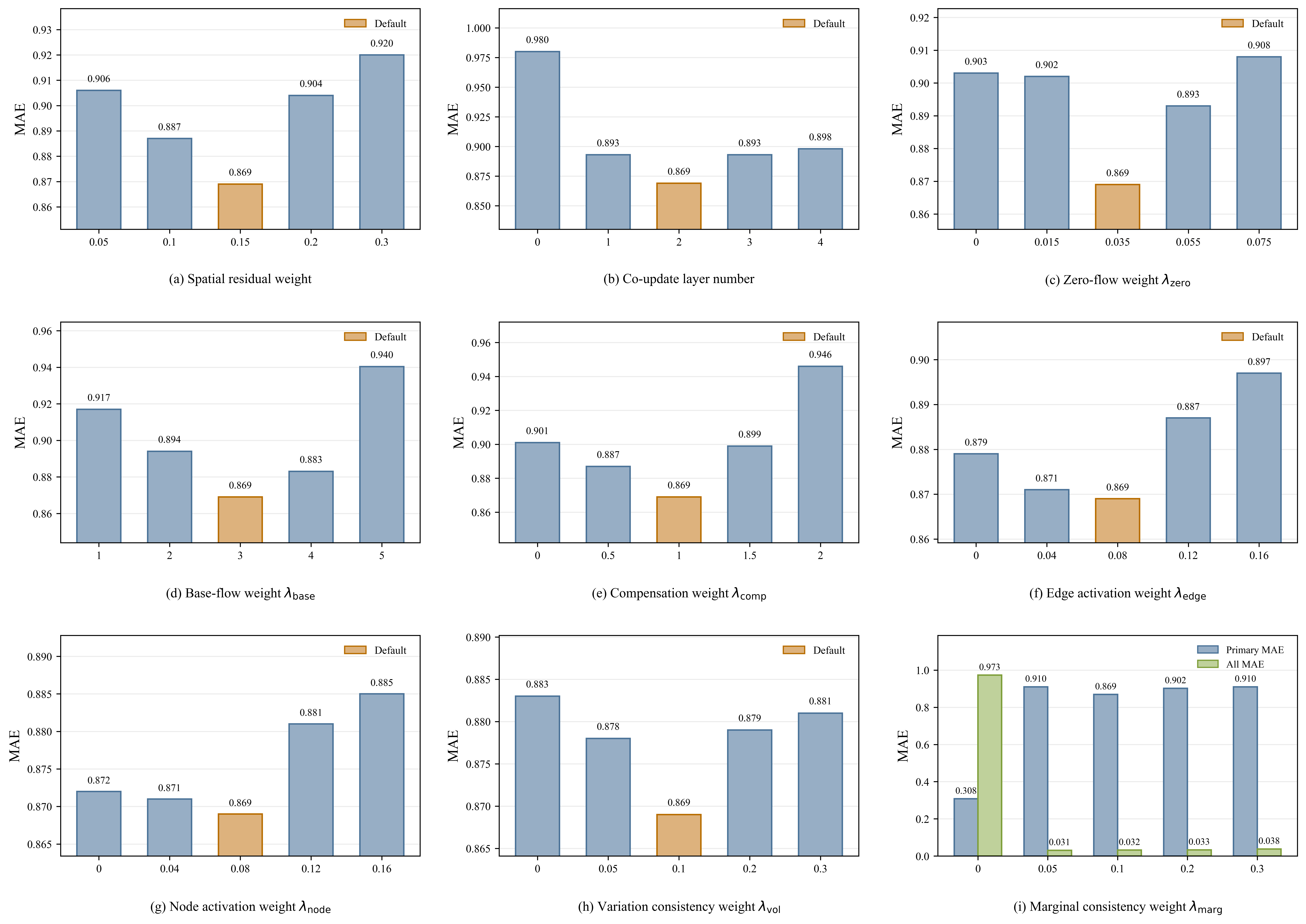}
\caption{Parameter sensitivity analysis of SAGMTL on the Beijing dataset. 
Panels (a)--(i) report the effects of the spatial residual weight, the number of co-update layers, the zero-flow loss weight, the base-flow loss weight, the compensation loss weight, the edge activity loss weight, the node activity loss weight, the volatility consistency weight, and the marginal consistency weight, respectively. 
The orange bars denote the default hyperparameter settings used in the final model. 
For most parameters, the primary MAE on active OD edges is reported. 
In panel (i), both the primary MAE and all-edge MAE are shown to illustrate the effect of marginal consistency on global OD flow allocation.}
\label{fig:sensitivity}
\end{figure*}

To evaluate the sensitivity of SAGMTL to key hyperparameters, we analyze the effects of the spatial dependency enhancement weight $\lambda_p$, the number of node--edge collaborative update layers $L$, the structural auxiliary task weights $\lambda_{\mathrm{edge}}$ and $\lambda_{\mathrm{node}}$, the flow modeling weights $\lambda_{\mathrm{base}}$ and $\lambda_{\mathrm{comp}}$, and the constraint weights $\lambda_{\mathrm{zero}}$, $\lambda_{\mathrm{vol}}$, and $\lambda_{\mathrm{marg}}$. The results are shown in Fig.~\ref{fig:sensitivity}.

For spatial dependency modeling, increasing $\lambda_p$ from 0.05 to 0.15 reduces MAE from 0.906 to 0.869, indicating that an appropriate geographical distance prior improves spatial dependency learning. However, a larger $\lambda_p$ increases prediction error, suggesting that overly strong spatial constraints may limit the modeling of long-distance functional correlations and non-local OD interactions. Therefore, $\lambda_p=0.15$ provides a suitable balance between spatial priors and data-driven learning.

For the node--edge collaborative update depth, the model performs poorly when $L=0$, with an MAE of 0.980, showing the importance of information propagation between node states and OD interactions. The best performance is achieved when $L=2$, while deeper propagation leads to performance degradation due to possible over-smoothing. Thus, a two-layer collaborative update structure is adopted.

For structural auxiliary task weights, both $\lambda_{\mathrm{edge}}$ and $\lambda_{\mathrm{node}}$ improve performance as they increase from 0 and reach the optimum around 0.08. This demonstrates that edge and node activity prediction provide useful structural supervision for dynamic sparse OD modeling. When the weights become too large, MAE increases, indicating that excessive auxiliary supervision may weaken the optimization of the main flow regression task. Edge activity supervision brings slightly greater gains than node activity supervision, suggesting that edge-level connection states are more directly related to OD flows.

For flow modeling parameters, increasing $\lambda_{\mathrm{base}}$ from 1 to 3 improves MAE, but further increasing it degrades performance, showing that the main regression objective should be balanced with structural constraints. For the intensity compensation loss, the best performance is obtained at $\lambda_{\mathrm{comp}}=1$. Removing this term weakens the modeling of medium- and high-intensity active connections, while an overly large weight may overemphasize high-flow samples and reduce generalization.

For constraint weights, increasing $\lambda_{\mathrm{zero}}$ to 0.035 improves performance by suppressing false positive flows on inactive edges, but an excessive weight may suppress low-frequency and weak connections. The marginal consistency constraint achieves the best performance at $\lambda_{\mathrm{marg}}=0.1$, indicating that appropriate inflow--outflow consistency improves structural rationality. The volatility consistency loss also shows a first-decreasing-then-increasing trend, suggesting that moderate temporal variation constraints help capture dynamic trends, whereas overly strong constraints may suppress bursty fluctuations.

Overall, SAGMTL is relatively stable within reasonable hyperparameter ranges. Spatial priors, structural auxiliary tasks, and flow constraints all improve prediction performance when properly weighted, while overly strong constraints may interfere with the main flow prediction objective. Therefore, the final parameter settings are chosen to balance prediction accuracy and structural modeling capability.

\section{CONCLUSION AND FUTURE WORK}
To address the challenges of identifying connection states, modeling long-tailed connections, and insufficiently exploiting the coupling between node states and edge-level interactions in dynamic sparse OD demand prediction, this paper proposes a structure-aware graph multi-task learning framework, SAGMTL. The framework jointly models node activity states, OD connection states, and edge-level flow intensities, and captures the coupling between regional states and OD interactions through a node--edge collaborative update mechanism. Meanwhile, SAGMTL introduces a steady--dynamic flow decomposition mechanism and multiple structural constraints to enhance its capability in predicting dynamic sparse OD demand.

Experiments on three real-world urban mobility datasets from Beijing, Chengdu, and Nanjing show that SAGMTL achieves the best performance in terms of MAE and MAPE, while maintaining strong competitiveness in RMSE. Further ablation studies and parameter sensitivity analysis verify the effectiveness of the structural auxiliary tasks, spatial dependency enhancement, node--edge collaborative update, and steady--dynamic flow decomposition mechanism. These results demonstrate that explicitly modeling the relationships among connection states, regional states, and flow intensities can improve dynamic sparse OD demand prediction.

Future work will further incorporate external factors such as weather, holidays, and events to enhance the model's ability to capture complex travel demand variations. In addition, continuous-time dynamic graph modeling and adaptive structure learning will be explored to more precisely characterize the evolution of OD connections. The proposed framework can also be extended to larger-scale urban areas and multimodal transportation scenarios, thereby improving its practical value in urban traffic management and intelligent mobility services.

\vfill

\end{document}